\def\year{2021}\relax
\documentclass[letterpaper]{article} 
\usepackage{aaai21}  
\usepackage{times}  
\usepackage{helvet} 
\usepackage{courier}  
\usepackage[hyphens]{url}  
\usepackage{graphicx} 
\usepackage{graphicx}  
\usepackage{natbib}  
\usepackage{caption} 
\usepackage{mathtools}
\usepackage{amsmath}
\usepackage{amssymb}
\usepackage{tabularx}
\usepackage{cleveref}
\usepackage{xcolor, colortbl}
\usepackage{dsfont}
\usepackage{bm}
\usepackage[toc,page]{appendix}
\nocopyright
\newcommand{\pad}{\vskip -0.125in}
\newcommand{\padT}{\vskip -0.05in}

\title{Joint Variational Autoencoders for Recommendation with Implicit Feedback}
\author {       Bahare Askari,\textsuperscript{\rm 1}
        Jaroslaw Szlichta, \textsuperscript{\rm 1}
        Amirali Salehi-Abari \textsuperscript{\rm 1} \\
}
\affiliations {
    \textsuperscript{\rm 1} Ontario Tech University, Canada \\
 
    bahare.askarifiroozjayi@ontariotechu.ca, jarek@ontariotechu.ca, abari@ontariotechu.ca
}
\DeclarePairedDelimiterX{\infdivx}[2]{(}{)}{
  #1\;\delimsize|\delimsize|\;#2
}
\newcommand{\kld}[2]{\mathit{KL}\infdivx{#1}{#2}}
\newcommand{\lvae}{L_{\scriptscriptstyle \mathtt{VAE}}}
\newcommand{\ljova}{L_{\scriptscriptstyle \mathtt{JoVA}}}
\newcommand{\ljovah}{L_{\scriptscriptstyle \mathtt{JoVA-H}}}
\newcommand{\lh}{L_{\scriptscriptstyle \mathtt{H}}}
\newcommand{\bftheta}{\pmb{\theta}}
\newcommand{\bfpsi}{\pmb{\psi}}
\newcommand{\bfphi}{\pmb{\phi}}
\newcommand{\ind}[1]{\mathds{1}\left[#1\right]}
\definecolor{lgreen}{rgb}{0.92, 0.92, 0.92}
\begin{document}

\maketitle

\begin{abstract}
Variational Autoencoders (VAEs) have recently shown promising performance in collaborative filtering with implicit feedback. 
These existing recommendation models learn user representations to reconstruct or predict user preferences. We introduce joint variational autoencoders (JoVA), an ensemble of two VAEs, in which VAEs jointly learn both user and item representations and collectively reconstruct and predict user preferences. This design allows JoVA to capture user-user and item-item correlations simultaneously. By extending the objective function of JoVA with a hinge-based pairwise loss function (JoVA-Hinge), we further specialize it for top-k recommendation with implicit feedback. Our extensive experiments on several real-world datasets show that JoVA-Hinge outperforms a broad set of state-of-the-art collaborative filtering methods, under a variety of commonly-used metrics. Our empirical results also confirm the outperformance of JoVA-Hinge over existing methods for cold-start users with a limited number of training data.

\end{abstract}

\section{Introduction}

The information overload and abundance of choices on the Web have made recommendation systems indispensable in facilitating user decision-making and information navigation. Recommender systems provide personalized user experience by filtering relevant items (e.g., books, music, or movies) or information (e.g., news). Many efforts have been devoted to developing effective recommender systems and approaches \cite{aggarwal2016recommender,s+x+k+2009}.

\textit{Collaborative filtering (CF)}---a well-recognized approach in recommender systems---is based on the idea that users with similar revealed preferences might have similar preferences in the future \cite{s+x+k+2009}.  User preferences in CF techniques are in the form of either \textit{explicit feedback} (e.g., ratings, reviews, etc.) or \textit{implicit feedback} (e.g., browsing history, purchasing history, search patterns, etc.). While explicit feedback is more informative than its implicit alternative, it imposes more cognitive burden on users through their elicitation, is subject to noisy self-reporting \cite{Amatriain2009}, and suffers from interpersonal comparison or \textit{calibration} issues \cite{s_b+s_c+2012,s_h+l_v+2009}. In contrast, implicit feedback naturally originates from user behavior based on the assumption that a user's interaction with an item is a signal of his/her interest to the item. Compared to explicit feedback, implicit feedback is more easily collected and abundant as long as user-item interactions are observable. 

This abundance of implicit feedback has made collaborative filtering more intriguing at the cost of some practical challenges. The implicit feedback lacks negative examples as the absence of a user-item interaction is not necessarily indicative of user disinterest (e.g., the user is unaware of the item).  Also, the user-item interaction data for implicit feedback is large, yet severely \emph{sparse}. It is even more sparse than explicit feedback data as the unobserved user-item interactions are a mixture of both missing values and real negative feedback.  Many attempts have been made to address these challenges by deep learning \cite{s_z+l_y+a_s+y_t+2019}.

Deep neural networks, with their representation learning power, are effective in capturing non-linear and sparse user-item interactions for recommendation with implicit feedback. Multilayer perceptron (or feedforward) networks were (arguably) the first class of neural networks successfully applied for collaborative filtering \cite{h_c+l_k+j_h+2016,x_h+l_l+h_z+2017}. Also, there has been emerging interest in deploying the variants of autoencoders such as  classical \cite{z+z+w+j+2019}, denoising \cite{w+y+d+c+2016}, and variational \cite{l+x+s+j+2017,l+d+k+r+2018}. However, these solutions either do not capture uncertainty of the latent representations \cite{z+z+w+j+2019,w+y+d+c+2016}, or solely focus on latent representation of users \cite{l+x+s+j+2017,l+d+k+r+2018}. Our work intends to address these shortcomings. 

We present \emph{joint variational autoencoder (JoVA)}, an ensemble of two variational autoencoders (VAEs). The two VAEs jointly learn both user and item representations while modeling their uncertainty, and then collectively reconstruct and predict user preferences. This design allows JoVA to capture user-user and item-item correlations simultaneously. We also introduce \emph{JoVA-Hinge}, a variant of JoVA, which extends the JoVA's objective function with a pairwise ranking loss to further specialize it for top-k recommendation with implicit feedback. Through extensive experiments over three real-world datasets, we show the accuracy improvements of our proposed solutions over a variety of state-of-the-art methods, under different metrics. Our JoVA-Hinge significantly outperforms other methods in the sparse datasets (up to 34\% accuracy improvement). Our experiments also demonstrate the outperformance of JoVA-Hinge across all users with varying numbers of training data (including cold-start users). Our findings confirm that the ensemble of VAEs equipped with pairwise loss improves recommendation with implicit feedback. Our proposed methods can potentially enhance other applications besides recommender systems.

\section{Related Work}
We review the related work on CF with implicit feedback.
\subsection{Implicit Feedback Recommendation}
Implicit feedback (e.g., clicking, browsing, or purchasing history) is a rich source of user preferences 
for recommender systems. This has motivated the development of various collaborative filtering methods, which exploit implicit feedback for effective recommendation \cite{y_h+y_k+c_v+2008,he2016fast}. 
The key developments are in either designing new models for capturing user-item interactions or novel objective functions for model learning.

\emph{Matrix factorization (MF)} and its variants \cite{koren2008factorization,r_s+a_m+2008} are among the successful classical models and techniques deployed for CF. In MF, users and items are represented in a shared low-dimensional latent space. Then, a user's interaction with an item is computed by the inner product of their latent vectors. 

Several well-known methods have formulated the recommendation task as a ranking problem and/or optimize ranking losses. \emph{Bayesian personalized ranking (BPR)} \cite{s_r+c_f+z_g+l_s+2012}, by assuming that users prefer an interacted item to an uninteracted one, minimizes its pairwise ranking loss for model learning. Its pairwise loss has been still deployed in many state-of-the-art methods; see for example  \cite{he2016fast,y+f+g+g+2016}. \textit{CofiRank} \cite{weimer2008cofi} directly optimizes ranking metrics by fitting a maximum margin matrix factorization model \cite{srebro2005maximum}. \textit{EigenRank} \cite{liu2008eigenrank} optimizes a function with Kendall rank correlation. RankALS \cite{takacs2012alternating} minimizes a ranking objective function with the \textit{alternating least squares (ALS)} method. 

These classical methods, despite their success in the recommendation, suffer from some limitations: (i) they fail to capture non-linear relationships between users and items; (ii) they cannot learn diverse user preferences as they treat each dimension of the latent feature space in the same way; and (iii) they have poor performance on sparse datasets. 

\subsection{Deep Recommendation}
Recently, deep learning has been promising for the recommendation tasks \cite{s_z+l_y+a_s+y_t+2019} by capturing more enriched representations for users, items, and their interactions. Neural collaborative filtering (NCF) \cite{x_h+l_l+h_z+2017} uses a multi-layer perceptron (MLP) to learn the user-item interaction function, and can be viewed as the generalization of matrix factorization. The Wide \& Deep model \cite{h_c+l_k+j_h+2016}---an app recommender for Google play---consists of two components. The wide component is a generalized linear model that handles cross-product features, whereas the deep component extracts nonlinear relations among features. The model learns item features through a feed-forward neural network with embeddings.
Another example of neural network-based recommender systems with implicit feedback is a visual Bayesian personalized ranking (VBPR) \cite{he2016vbpr}, which is an extension of the Bayesian personalized ranking with visual features. 

Of the most relevant to our work are recommender systems built based on autoencoders or their variations. \textit{Collaborative deep ranking (CDR)} \cite{ying2016collaborative} jointly implements representation learning and collaborative ranking by employing stacked denoising autoencoders. \textit{Joint collaborative autoencoder (JCA)} \cite{z+z+w+j+2019} deploys two separate classical autoencoders jointly optimized only by a hinge loss function for capturing user-user and item-item correlations. The proposed mini-batch optimization algorithm allows optimizing JCA without loading the entire rating matrix. Mult-VAE \cite{l+d+k+r+2018} is a collaborative filtering model for implicit feedback based on variational autoencoders. Mult-VAE uses a multinomial log-likelihood instead of the Gaussian likelihood. This work also proposed a regularization hyperparameter to control the trade-off between the reconstruction loss and the Kullback-Leibler (KL) loss in the objective function. Recently, RecVAE \cite{s+i+a+a+t+2019} proposed a new approach to optimizing this hyperparameter.

Our work is closely related to both JCA and Mult-VAE, as we build on the strengths of these two.  While JCA is jointly optimizing two classical autoencoders, it does not capture the uncertainity of latent representations, and consequently does not benefit from  representation power of variational autoencoders. To address this, we deploy two separate variational autoencoders and jointly optimized them by our proposed loss function. Our loss function, by taking into account two variational autoencoders' losses and a pairwise ranking loss, well tunes our deep learning models for recommendation with implicit feedback. Our proposed work, while differentiating from both JCA and Mult-VAE with regards to both architecture and loss function, can be viewed as the powerful generalization or extension of these two.

\section{Preliminaries}
Our goal is to provide personalized item recommendations with the presence of implicit feedback. In this section,  we formally define our problem, and describe VAE, which serves as a building block for our proposed model.
\subsection{Recommendation with Implicit Feedback}
We assume that a set of $n$ users $U$ can interact with the set of $m$ items $I$ (e.g., users click ads, purchase products, watch movies, or listen to musics). We consider user-item interactions are binary (e.g., a user has watched a specific movie or not), and represent them with the user \emph{implicit feedback matrix} $\mathbf{R} \in \{0, 1\}^{m\times n}$, where $\mathbf{R}_{ui}=1$, if the interaction of user $u$ with item $i$ is observed. As each column (or row) of the matrix corresponds to a specific item (or user), we let $\mathbf{R}_u$ and $\mathbf{R}^T_i$ denote the user $u$'s and item $i$'s interaction vector, respectively.  We also let  $I_u^+ = \{i \in I \mbox{ } | \mbox{ } \mathbf{R}_{ui} = 1\}$ denote a set of items that user $u$ has interacted with, and $I_u^- = I \setminus I_u^+$ be a set of items that user $u$ has not yet interacted with. 

Our goal in top-k recommendation is to suggest $k$ most preferred (or likely) items to user $u$ from $I_u^-$. To achieve this goal, we predict the likelihood of interaction between user $u$ and $I_u^-$ (or preference of user $u$ over $I_u^-$), and then select a rank-list of $k$ items with the highest prediction score to recommend to user $u$.  Our learning task is to find a \textit{scoring (or likelihood) function} $f$ that predicts an \emph{interaction score} $\hat{r}_{ui}$ for each user $u$ and an unobserved item $i \in I_u^-$. If $\hat{r}_{ui} \in [0,1]$, it can be interpreted as the predicted likelihood of user $u$'s interaction with item $i$. The function $f$ is formulated as $\hat{r}_{ui}$ = $f(u,i|\boldsymbol{\theta})$, where $\boldsymbol{\theta}$ denotes the model parameters. 

Most of model-based CF methods \cite{s+x+k+2009} differentiate from each other on the scoring function $f$ formulation or the objective functions used for parameter learning. There are various formulation of the function $f$, ranging from deep networks \cite{s_z+l_y+a_s+y_t+2019} to matrix factorization \cite{koren2008factorization}. In general, the objective functions fall into two categories. \emph{Pointwise loss} \cite{x_h+l_l+h_z+2017,y_h+y_k+c_v+2008}, by assuming an unobserved user-item interaction as a negative example, minimizes the error (or distance) between predicted interaction score $\hat{r}_{ui}$ and its actual value $r_{ui}$. In contrast to pointwise loss, \emph{pairwise loss} \cite{s_r+c_f+z_g+l_s+2012,he2016vbpr} directly optimizes the ranking of the user-item interaction while assuming that users prefer observed items to unobserved items. 

\subsection{Variational Autoencoder (VAE)}
Our model uses the variational autoencoder (VAE) \cite{d+p+w+m+2014} as a building block. VAE is a deep generative model for learning complex distributions. Each VAE, similar to classical autoencoders, consists of encoder and decoder networks. The encoder first encodes the inputs to latent representations, and then the decoder reconstructs the original inputs from latent representations. However, the VAE differentiates from classical autoencoders by encoding an input as a distribution over latent representations (rather than a single point). This choice of probabilistic representation not only makes VAE a generative model, but also reduces overfitting by forcing smoother latent representation transitions.

The encoder network of VAE encodes the input $\mathbf{x}$ to a d-dimensional latent representation $\mathbf{z}$, which is a multivariate random variable with a prior distribution $p(\mathbf{z})$.\footnote{The common practice is to assume that $p(\mathbf{z})$ is a standard multivariate normal distribution: $\mathbf{z} \sim \mathcal{N}(\mathbf{0},\mathbf{I})$.} One can view the encoder as the posterior distribution $p_{\bfphi}(\mathbf{z}|\mathbf{x})$ parametrized by ${\bfphi}$. Since this posterior distribution is intractable, it is usually approximated by variational distribution \cite{b+d+k+2017}:

\begin{equation}
    q_{\bfphi}(\mathbf{z}|\mathbf{x}) = \mathcal{N}(\mu_{\bfphi}(\mathbf{x}),\sigma_{\bfphi}^2(\mathbf{x})\mathbf{I}),
\end{equation}  
where two multivariate functions $\mu_{\bfphi}(\mathbf{x})$ and $\sigma_{\bfphi}(\mathbf{x})$ map the input $\mathbf{\mathbf{x}}$ to the mean and standard deviation vectors, respectively. In VAE, $\mu_{\bfphi}(\mathbf{x})$ and $\sigma_{\bfphi}(\mathbf{x})$ are jointly formulated by the \emph{inference network}  $f_{\bfphi}(\mathbf{x}) = [\mu_{\bfphi}(\mathbf{x}), \sigma_{\bfphi}(\mathbf{x})]$. 

The decoder network $p_{\bfpsi}(\mathbf{x}|\mathbf{z})$, also known as \emph{generative network}, takes $\mathbf{z}$ and outputs the probability distribution over (reconstructed) input data $\mathbf{x}$. Putting together the encoder and decoder networks, one can lower bound the log-likelihood of the input $\mathbf{x}$ by

\begin{align*}
    \log p(\mathbf{x})&\geq \int q_{\bfphi}(\mathbf{z}|\mathbf{x})  \log \frac{p_{\bfpsi}(\mathbf{x}|\mathbf{z}) p(\mathbf{z})}{q_{\bfphi}(\mathbf{z}|\mathbf{x})}dz\\
    &=E_{q_{{\bfphi}}(\mathbf{z}|\mathbf{x})}\left[\log p_{\bfpsi}(\mathbf{x}|\mathbf{z})\right] - \kld{q_{\bfphi}(\mathbf{z}|\mathbf{x})}{p(\mathbf{z})},
\end{align*}
where $\mathit{KL}$ is Kullback-Leibler divergence distance measuring the difference between the distribution $q_{\bfphi}(\mathbf{z}|\mathbf{x})$ and the unit Gaussian distribution $p(\mathbf{z})$. This lower bound, known as \emph{evidence lower bound (ELBO)}, is maximized for learning the parameters of encoder and decoder, ${\bfphi}$ and $\bfpsi$, respectively. Equivalently, for learning VAE parameters, one can minimize the negation of the ELBO as a loss function (see Eq.~\ref{eq:vae_loss}) by stochastic gradient decent with the reparameterization trick \cite{d+p+w+m+2014}.

\begin{equation}
\lvae(\mathbf{x}|\bftheta) =  - E_{q_{\bfphi}(\mathbf{z}|\mathbf{x})}[\log p_{\bfpsi} (\mathbf{x}|\mathbf{z})] + \kld{q_{\bfphi}(\mathbf{z}|\mathbf{x})}{p(\mathbf{z})},
\label{eq:vae_loss}
\end{equation}
where $\bftheta=[\bfpsi,\bfphi]$.  This loss function can be viewed as a linear combination of \emph{reconstruction loss} and KL divergence, which serves as a regularization term. In this light, recent research  \cite{l+d+k+r+2018,s+i+a+a+t+2019} has introduced regularization hyperparameter $\alpha$ for controlling the trade-off between regularization term and reconstruction loss:

\begin{equation}
\lvae(\mathbf{x}|\bftheta, \alpha) =  - E_{q_{\bfphi}}[\log p_{\bfpsi} (\mathbf{x}|\mathbf{z})] + \alpha \kld{q_{\bfphi}(\mathbf{z}|\mathbf{x})}{p(\mathbf{z})},
\label{eq:vae_loss-beta}
\end{equation}

As our input data $\mathbf{x}$ is a binary vector (i.e., implicit feedback), we consider logistic likelihood for the output of the VAE decoder. Defining $f_{\bfpsi}(\mathbf{z}) = [o_i]$ as the output of generative function of the decoder, the logistic log-likelihood for input $\mathbf{x}$ is  

\begin{equation}
   \log p_{\bfpsi}(\mathbf{x}|\mathbf{z}) = \sum_i x_i \log \sigma (o_i) + (1 - x_i) (1 - \sigma (o_i)).
   \label{eq:logliklihood}
\end{equation}
Here, $\sigma (x) = 1/(1+\exp(-x))$ is the logistic function. This logistic likelihood renders the reconstruction loss to the cross-entropy loss.  

\section{Joint Variational Autonecoder (JoVA)}

\begin{figure}[t]
  \centering
  \includegraphics[width=\linewidth]{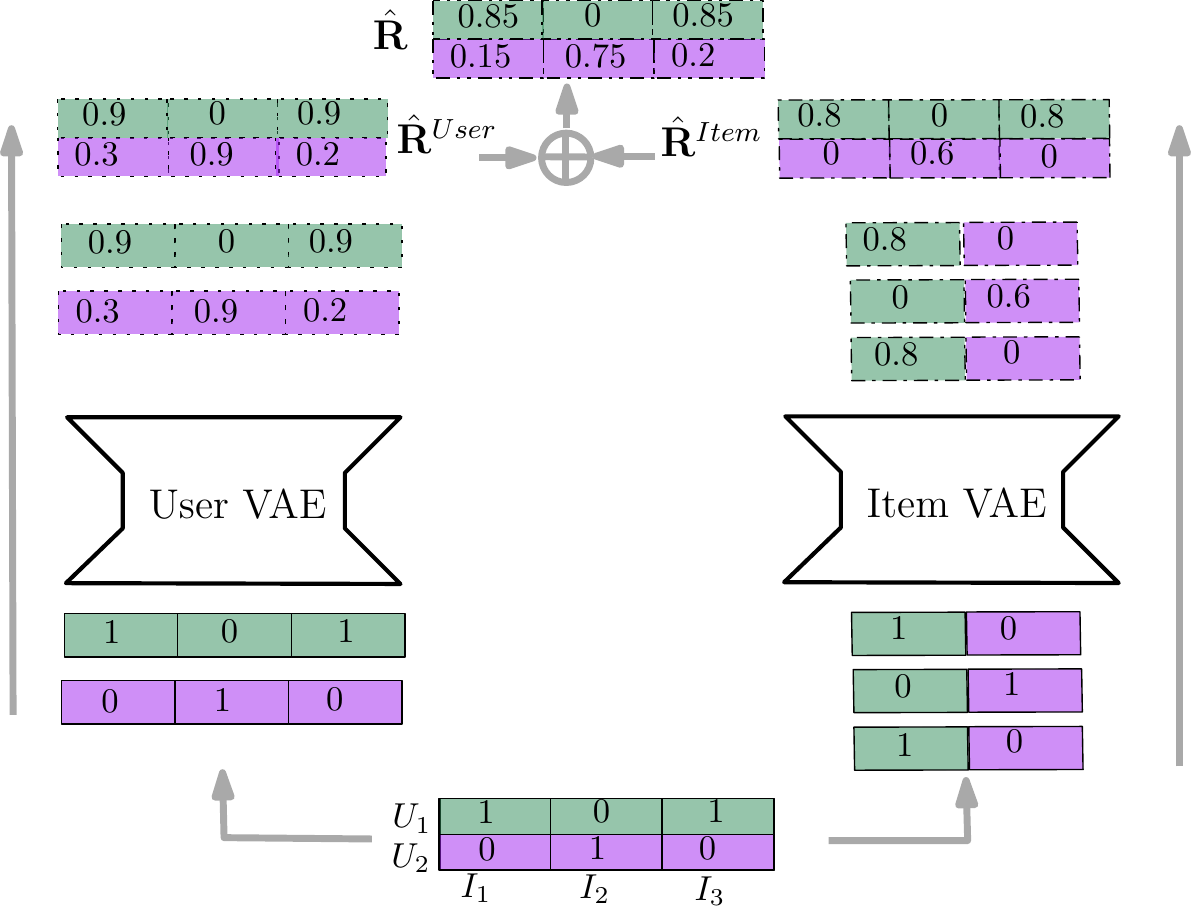}
  \caption{Illustration of the JoVA model. User and item VAEs recover the input matrix independently. The final output is the average of these two reconstructed matrices.}
  \label{fig:jova-model}
  \pad
\end{figure}

We here detail our proposed \emph{joint variational autoencoder (JoVA)} framework and its variant \emph{JoVA-Hinge} for top-k recommendation with implicit feedback. We first discuss the model architecture of JoVA and then discuss various objectives functions used for parameter learning. 

\subsection{Model}\label{sec:JVA}
Our model consists of two separate variational autoencoders of \emph{user VAE} and \emph{item VAE} (see Figure \ref{fig:jova-model}). Given the implicit feedback matrix $\mathbf{R}$, the user VAE aims to reconstruct the matrix row-by-row, whereas the item VAE reconstructs it column-by-column. In other words, user VAE takes and reconstruct each user vector  $\mathbf{R}_u$ (i.e., a row of the matrix). Similarly, item VAE takes and reconstruct each item vector $\mathbf{R}^T_i$ (i.e., a column of the matrix). These two VAEs independently and simultaneously complete the implicit feedback matrix. The final output of our model is the average of two predicted implicit matrices:
\begin{equation}
    \hat{\mathbf{R}} = \frac{1}{2} (\hat{\mathbf{R}}^{user} + \hat{\mathbf{R}}^{item}),
\end{equation}
where $\hat{\mathbf{R}}^{user}$ and $\hat{\mathbf{R}}^{item}$ are implicit matrices predicted (or completed) by the user VAE and the item VAE, respectively. We note that $\hat{\mathbf{R}} \in [0,1]^{m\times n}$, where each $\hat{r}_{ui}$ represents the predicted likelihood that user $u$ interacts with item $i$. This natural probabilistic interpretation originates from our choice of logistic likelihood for the output of VAEs (see Eq.~\ref{eq:logliklihood}).  
The parameters of user VAE and item VAE are learned jointly with a joint objective function (see Sec. \ref{sec:obj}). 

JoVA model is designed carefully to capture both user-user and item-item correlations. The item VAE encodes similar items close to each other in its latent representations to preserve their correlations, while the user VAE does the same for similar users. The joint optimization of these two VAEs helps their fine-tune calibration, so that they can complement each other in their predictions. The item and user VAEs together can learn complementary information from user-item interactions beyond what each could separately learn. This richer learning is a valuable asset for sparse datasets, as confirmed by our experiments in Sec.~\ref{sec:exp}.

One can readily observe the connections between JoVA and ensemble learning. Similar to ensemble learning, JoVA combines user VAE and item VAE into one learning framework for the final prediction.  From this perspective, each VAE independently predicts the rating matrices, and then the final prediction is the aggregation of VAEs' predictions. The aggregation in JoVA is with unweighted averaging, which is shown to be a reliable choice as an aggregation method in the ensemble of deep learning models \cite{j+c+b+a+2018}. This unweighted averaging can easily be extended to the weighed averaging at the cost of tuning more hyper-parameters for each dataset, but with the promise of increased accuracy.\footnote{We have confirmed this in some experiments not reported in this paper.} Averaging user VAE and item VAE predictions can reduce the expected variance of neural network models and consequently, the risk of overfitting, thus, improving model accuracy. 

\subsection{Objective functions}\label{sec:obj}
To learn model parameters of JoVA, we consider two variants of loss functions. One naturally arises from the combination of two user and item variatianal autoencoders:
\begin{equation}
    \ljova(\mathbf{R}|\bftheta,\alpha) = \sum_{u\in U} \lvae(\mathbf{R}_u|\bftheta_{U},\alpha) + \sum_{i \in I} \lvae( \mathbf{R}^T_i|\bftheta_{I}, \alpha)
\end{equation}
Here, $\bftheta_{U}$ and $\bftheta_{I}$ represent the model parameters of user and item VAEs repsectively, and $\lvae$ is computed by Eq.~\ref{eq:vae_loss-beta} with the logistic likelihood of Eq.~\ref{eq:logliklihood}.

To further specialize JoVA model for the top-k recommendation, we incorporate a pairwise ranking loss in its joint loss function. Specifically, we introduce \emph{JoVA-Hinge (JoVA-H)} loss function: 
\begin{equation}
    \ljovah(\mathbf{R}|\bftheta,\alpha, \beta , \lambda) = \ljova(\mathbf{R}|\bftheta,\alpha) +
    \beta \lh(\mathbf{R}|\bftheta,\lambda), 
    \label{eq:jova-hinge}
\end{equation}
where 
$$
\lh(\mathbf{R}|\bftheta, \lambda)  = \sum_{u\in U}\sum_{i\in I^+_u} \sum_{j\in I^-_u} \max (0, \hat{r}_{uj} - \hat{r}_{ui} +\lambda)
$$
is \emph{hinge loss function}, wildly and successfully used as a pairwise ranking loss \cite{z+z+w+j+2019,w+j+b+s+2011,y+t+m+t+2016}. Here,  $\hat{r}_{ui}$ is the predicted interaction score (or likelihood) of user u for item i, and $\lambda$ is the margin hyperparameter for the hinge loss. The hinge loss is built upon the assumption that user $u$ prefers his interacted item  $i \in I^+_u$  over an uninteracted item (or negative example)  $j \in I^-_u$ with the margin of at least $\lambda$.\footnote{In practice,  the hinge loss is usually computed over a sample of negative examples.} We have introduced the hyperparameter  $\beta$ for controlling the influence of hinge loss to the JoVA's objective function.

\section{Experiments}\label{sec:exp}
Our empirical experiments intend to assess the effectiveness of our proposed methods JoVA and JoVA-Hinge for top-k recommendation with implicit feedback. We compare the accuracy of our methods (under various evaluation metrics) with an extensive set of state-of-the-art methods on three real-world datasets. We further study the effectiveness of our methods in handling cold-start users. 
The source code is available on \footnote{\url{https://github.com/bahareAskari/JoVA-Hinge.git}}

\begin{table}[tb]
\setlength\tabcolsep{5pt}
\begin{center}
\begin{tabular}{|l|l|l|l|l|}
\hline
\textbf{Dataset} &\textbf{\#User}& \textbf{\#Item}&\textbf{\#Interaction}&\textbf{Sparsity}\\
\hline 
\hline
MovieLens&6,027&3,062 &574,026 &96.89\%\\
Yelp&12,705 &9,245 &318,314 &99.73\%\\
Pinterest&55,187&9,911&1,500,806&99.73\%\\
\hline
\end{tabular}
\padT
\caption{The summary statistics of our tested datasets.}
\label{tab:PPer}
\end{center}
\pad
\end{table}

\subsection{Evaluation Datasets}
We report results obtained on three real-world datasets: MovieLens-1M (ML1M)\footnote{\url{http://files.grouplens.org/datasets/movielens/ml-1m.zip}.}, Pinterest\footnote{\url{https://sites.google.com/site/xueatalphabeta/academic-projects}}, and  Yelp\footnote{\url{https://www.yelp.com/dataset/challenge}.}.  
Pinterest is a dataset with implicit feedback in which the interaction of a user with an image (i.e., item) is 1, if the user has pinned the image to their own board.  Following the previous work \cite{x_h+l_l+h_z+2017}, we kept only users with at least 20 interactions (pins).
ML1M and Yelp originally include five-star user-item ratings.  A user-item rating was converted to 1, if it is greater than or equal to 4 and to 0 otherwise. This method for converting explicit feedback to implicit feedback is a common practice, for example, see  \cite{l+h+w+j+2019,z+z+w+j+2019,l+d+k+r+2018}. Table \ref{tab:PPer}  provides the summary statistics of these datasets after pre-processing. For each dataset, we randomly selected 80\% of all user-item interactions as the training set and equally split the remaining 20\% into testing and validation sets.

\begin{figure*}[tb]
    \centering
    \begin{tabular}{@{\hspace{-7pt}}c@{\hspace{-1pt}}c@{\hspace{-1pt}}c}
       \includegraphics[width=0.33\textwidth]{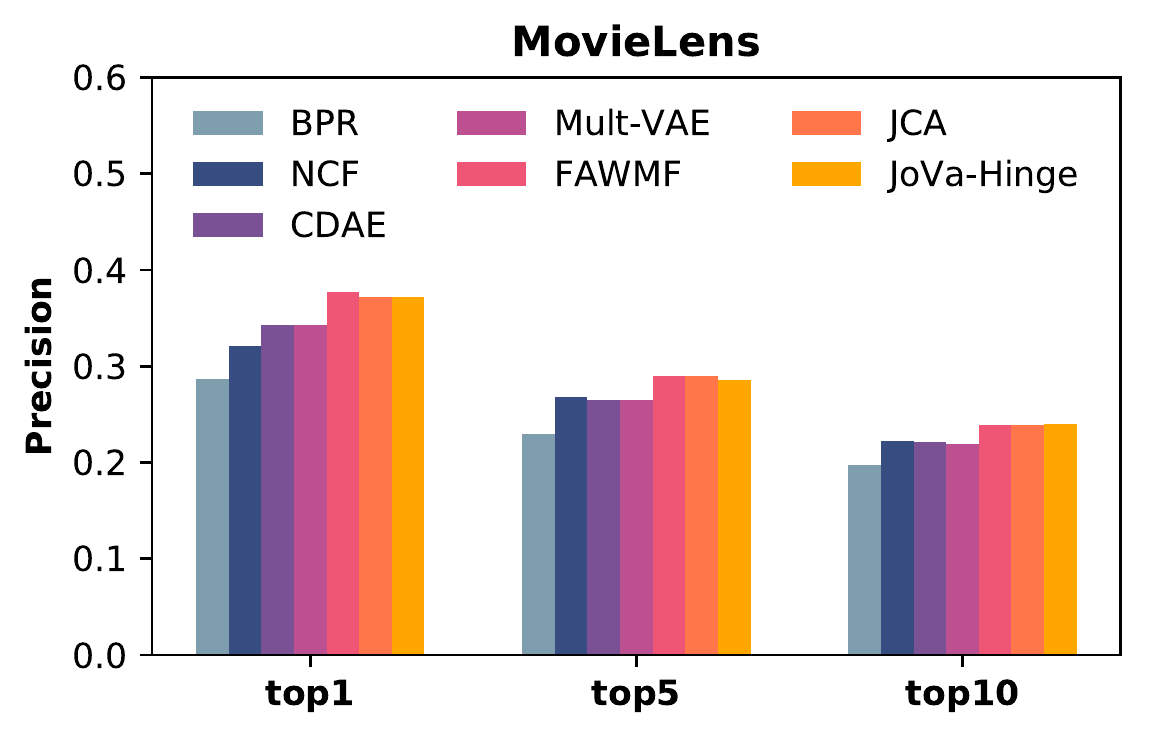}  &
       \includegraphics[width=0.33\textwidth]{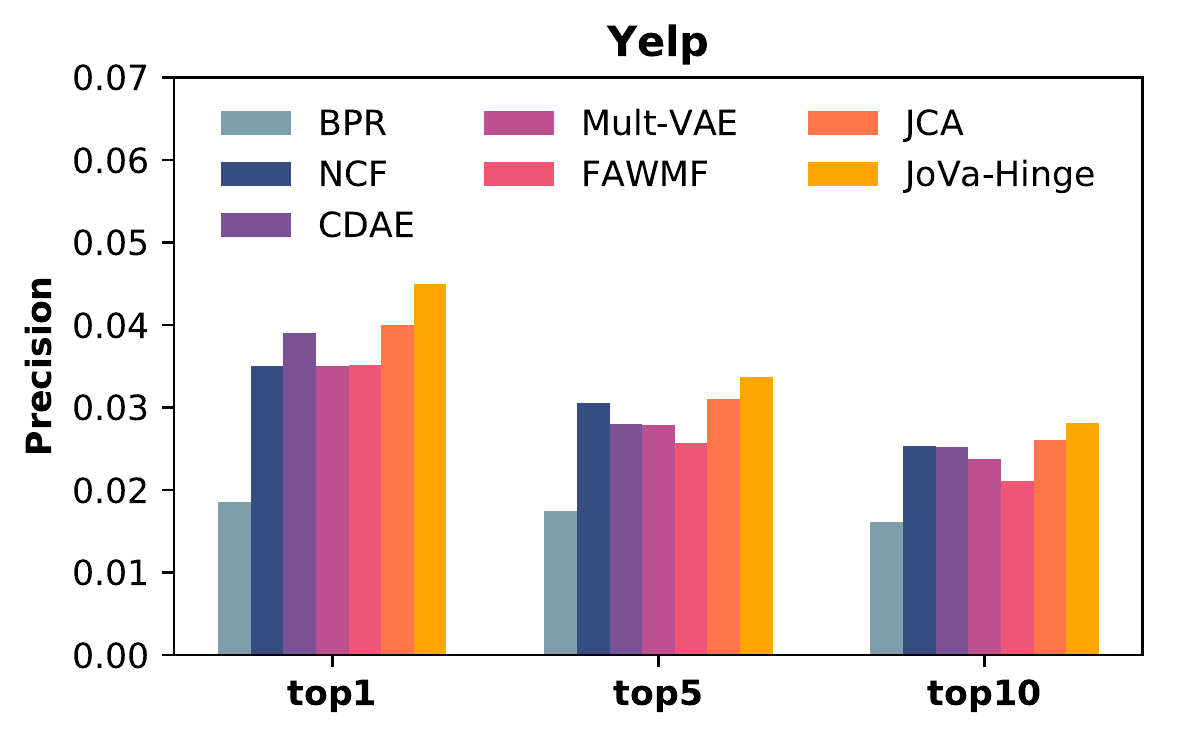} &
       \includegraphics[width=0.33\textwidth]{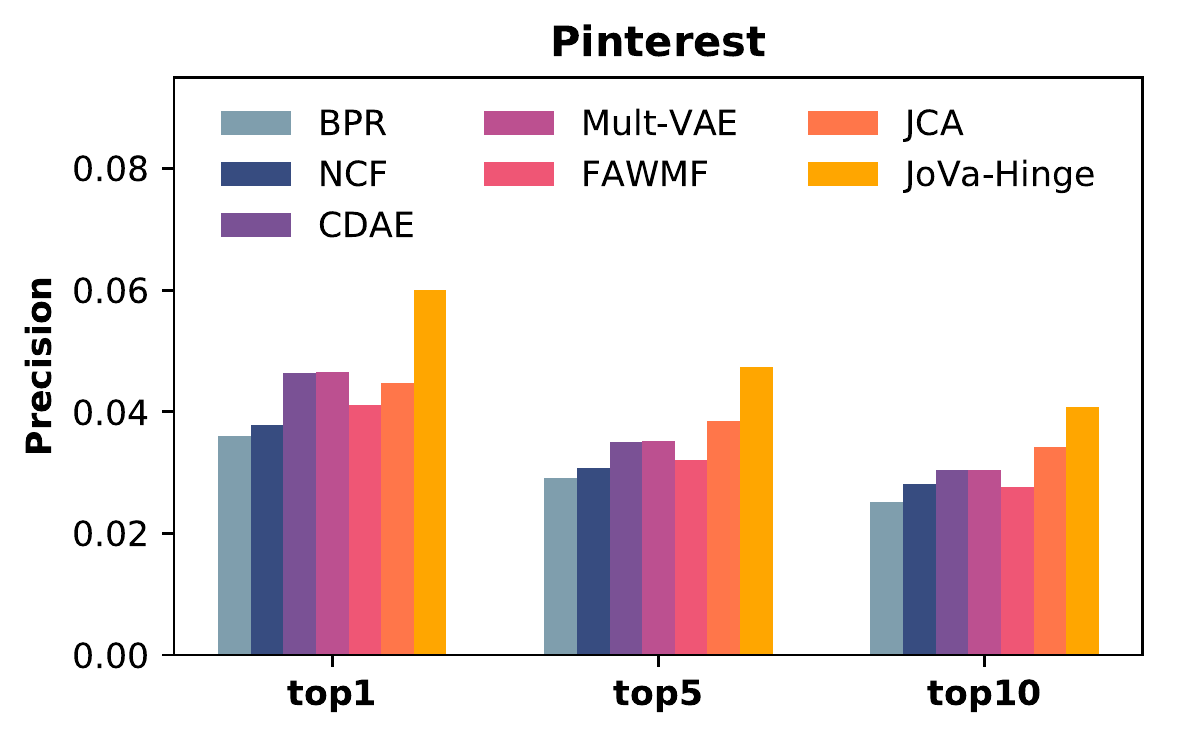}\\
       (a) Precision, MovieLens. &(b) Precision, Yelp. & (c) Precision, Pinterest.\\
       \includegraphics[width=0.33\textwidth]{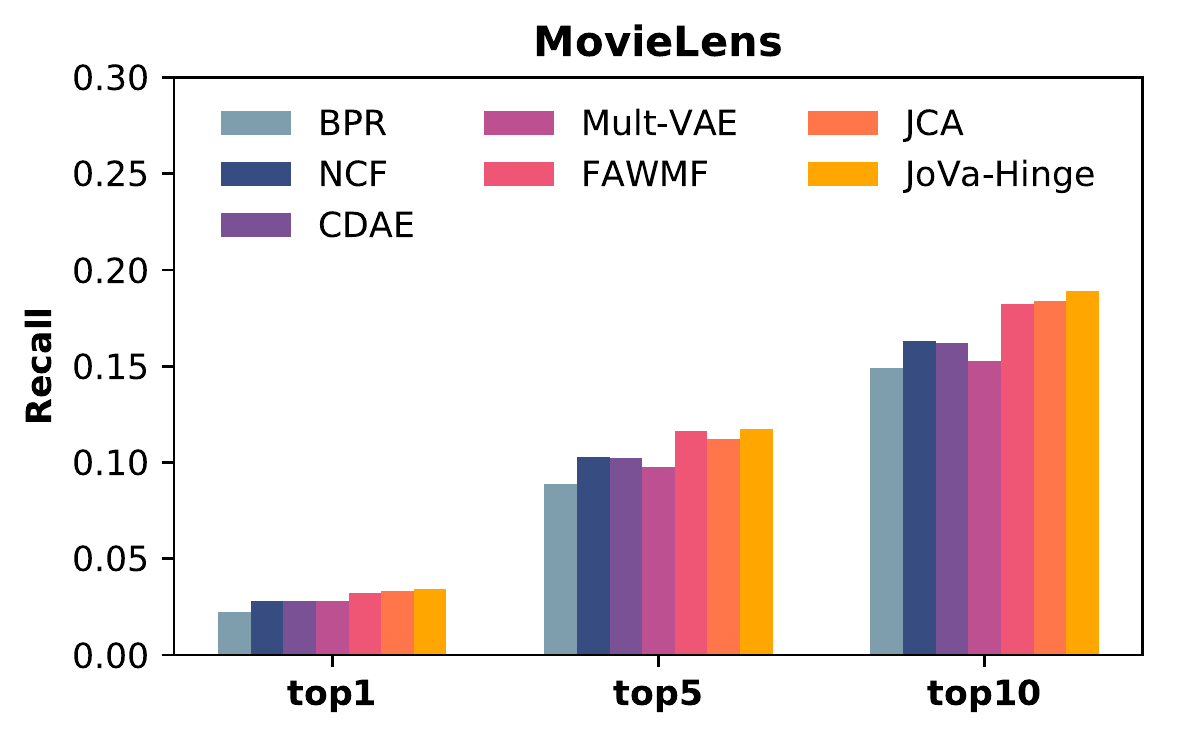}  &
       \includegraphics[width=0.33\textwidth]{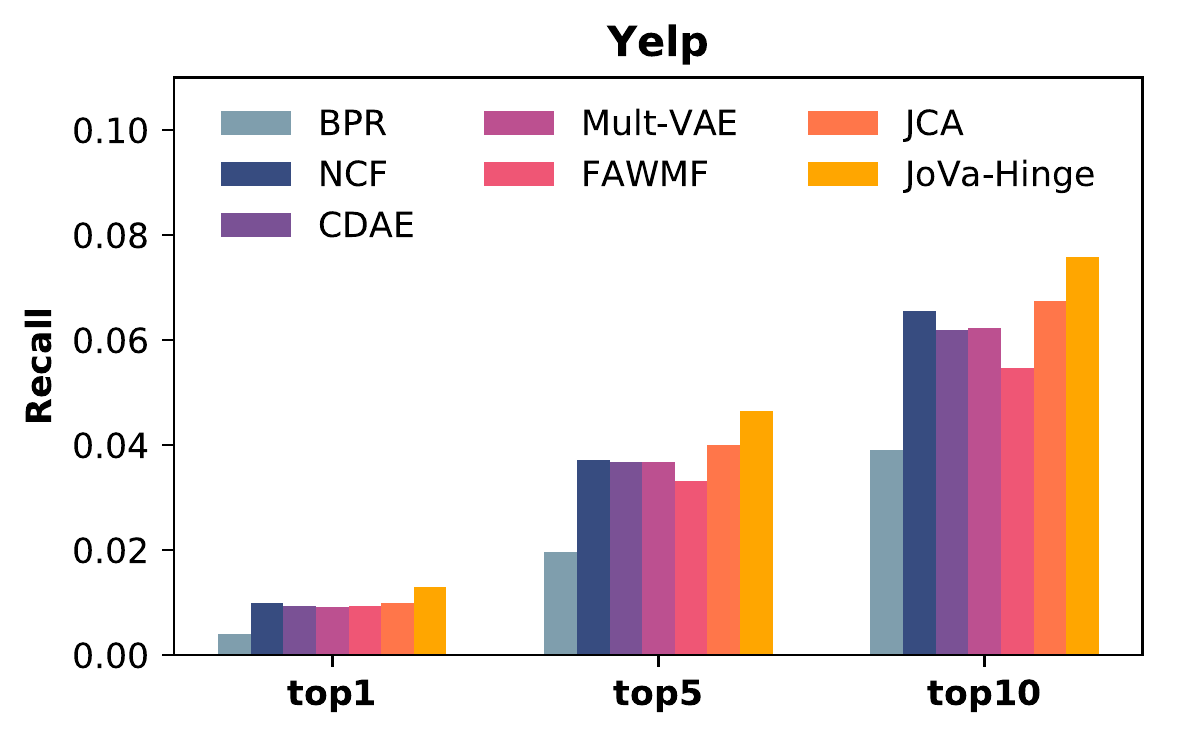} &
       \includegraphics[width=0.33\textwidth]{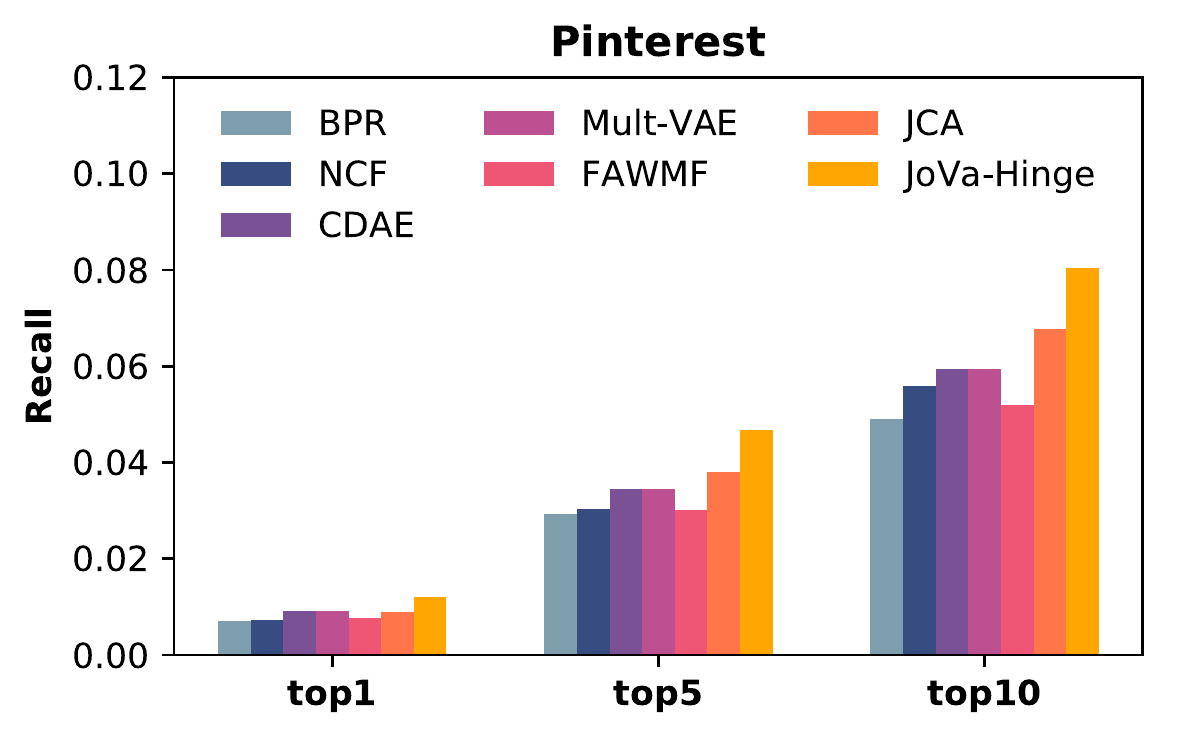}\\
       (d) Recall, MovieLens. & (e) Recall, Yelp. & (f) Recall, Pinterest.\\
    \end{tabular}
    \padT
    \caption{Precision (a--c) and Recall (d--f) for all methods and three datasets of MovieLens, Yelp, and Pinterest.}
    \padT
    \label{fig:pre-rec-all}
\end{figure*}

\begin{table*}[tb]
\small
\setlength\tabcolsep{2.34pt}
\begin{center}
\begin{tabular}{|l|c|c|c|c|c|c|c|c|c|c|c|c|c|c|c|c|c|c|}
\cline{2-19}
\multicolumn{1}{c|}{}&\multicolumn{6}{c|}{\textbf{ML1M}}& \multicolumn{6}{c|}{\textbf{Yelp}}&\multicolumn{6}{c|}{\textbf{Pinterest}}\\ 
\cline{2-19}
\multicolumn{1}{c|}{}&\multicolumn{3}{c|}{\textbf{F1-score}}& \multicolumn{3}{c|}{\textbf{NDCG}}& \multicolumn{3}{c|}{\textbf{F1-score}}& \multicolumn{3}{c|}{\textbf{NDCG}}& \multicolumn{3}{c|}{\textbf{F1-score}}& \multicolumn{3}{c|}{\textbf{NDCG}}\\ 
\cline{2-19}
\multicolumn{1}{c|}{}&\textbf{@1} & \textbf{@5}& \textbf{@10}& \textbf{@1}& \textbf{@5}&\textbf{@10}&\textbf{@1}&\textbf{@5}&\textbf{@10}&\textbf{@1}& \textbf{@5}& \textbf{@10}& \textbf{@1}& \textbf{@5}& \textbf{@10}& \textbf{@1}& \textbf{@5}& \textbf{@10}                  \\ \hline
\textbf{BPR}&.0410&.1285&.1698&.2843&.2549&.2434&.0065&.0180&.0219&.0166&.0223&.0301&.0120&.0292&.0333&.0328&.0312&.0414\\ \hline
\textbf{NCF}&.0513&.1487&.1883&.2955&.2727&.2709&.0153&.0325&.0350&.0367&.0392&.0497&.0123&.0306&.0375&.0375&.0348&.0479\\ \hline
\textbf{CDAE}&.0518&.1474&.1873&.3428&.2896&.2728&.0159&.0315&.0356&.0378&.0390&.0471&.0154&.0349&.0401&.0415&.0387&.0506\\ \hline
\textbf{Mult-VAE}&.0518&.1420&.1801&.3428&.2886&.2695&.0148&.0317&.0344&.0350&.0381&.0465&.0153&.0349&.0402&.0466&.0397&.0504\\ \hline
\textbf{FAWMF}&.0595&\textbf{.1661}&.2068&\textbf{.3775}&\textbf{.3176}&\textbf{.2991}&.0152&.0290&.0305&.0358&.0358&.0425&.0131&.0310&.0360&.0416&.0359&.0450\\ \hline
\textbf{JCA} &\textbf{.0602}&.1634                 &  \textbf{.2080}&.3699&.3125&.2976&\textbf{.0160}&\textbf{.0350}&\textbf{.0376}&\textbf{.0405}&\textbf{.0440}&\textbf{.0537}&\textbf{.0150}&\textbf{.0383}&\textbf{.0456}&\textbf{.0448}&\textbf{.0424}&\textbf{.0557}\\ \hline
\textbf{JoVA-H}&\textbf{.0624}&\textbf{.1665}&\textbf{.2115}&
\textbf{.3718}&\textbf{.3143}&\textbf{.3013}&\textbf{.0201}&\textbf{.0391}&\textbf{.0401}&\textbf{.0449}&\textbf{.0483}&\textbf{.0581}&\textbf{.0200}&\textbf{.0471}&\textbf{.0542}&\textbf{.0604}& \textbf{.0532}&\textbf{.0678}\\ \hline\hline
\textbf{\% improve}&3.65&0.24&1.68&-1.53&-1.04&0.73&25.62&11.71&6.64&10.86&9.77&8.19&33.33&22.97& 18.85&34.82&25.47&21.72\\ \hline
\end{tabular}
\caption{Performance of the baselines and JoVA-Hinge on three datasets under F1@k and NDCG@k metrics. The results of JoVA-Hinge and the best baselines are in bold.}
\label{tab:my-table}
\end{center}
\pad
\end{table*}

\begin{table*}
\small
\setlength\tabcolsep{4.1pt}
\begin{center}
\begin{tabular}{|l|l|l|l|l|l|l|l|l|l|l|l|l|}

\cline{2-13}
\multicolumn{1}{c|}{} & \multicolumn{4}{c|}{\textbf{ML1M}}&\multicolumn{4}{c|}{\textbf{Yelp}}&\multicolumn{4}{c|}{\textbf{Pinterest}}\\
\hline
&\textbf{P@1}&\textbf{R@1}&\textbf{F1@1}&\textbf{NDCG@1}&\textbf{P@1}&\textbf{R@1}&\textbf{F1@1}&\textbf{NDCG@1}&\textbf{P@1}&\textbf{R@1}&\textbf{F1@1}&\textbf{NDCG@1}\\ \hline \textbf{JoVA}&\cellcolor{lgreen}{0.3730}&0.0329&0.0605&\cellcolor{lgreen}{0.3730}&0.0420&0.0120&0.0180&0.0433&0.0571&0.0113&0.0189&0.0571\\ 
\hline \textbf{JoVA-H}& {0.3718}& {\cellcolor{lgreen}{0.0340}}&{\cellcolor{lgreen}{0.0624}}&{0.3718}&{\cellcolor{lgreen}{0.0449}} & {\cellcolor{lgreen}{0.0130}}&{\cellcolor{lgreen}{0.0201}}&{\cellcolor{lgreen}{{0.0449}}}&{\cellcolor{lgreen}{0.0604}}&{\cellcolor{lgreen}{0.0120}}& {\cellcolor{lgreen}{0.0200}}&{\cellcolor{lgreen}{0.0604}}\\ 
\hline
\hline
&\textbf{P@5}&\textbf{R@5}&\textbf{F1@5}&\textbf{NDCG@5}&\textbf{P@5}&\textbf{R@5}&\textbf{F1@5}&\textbf{NDCG@5}&\textbf{P@5}&\textbf{R@5}&\textbf{F1@5}&\textbf{NDCG@5}\\ \hline
   \textbf{JoVA}&0.2845&0.1169&0.1657&0.3135&0.0320&0.0430&0.0360&0.0449&0.0464&0.0459&0.0461&0.0516\\ \hline
      \textbf{JoVA-H}&{\cellcolor{lgreen}{0.2853}}&{\cellcolor{lgreen}{0.1176}}&{\cellcolor{lgreen}{0.1665}}&{\cellcolor{lgreen}{0.3143}}&{\cellcolor{lgreen}{0.0337}}&{\cellcolor{lgreen}{0.0464}}&{\cellcolor{lgreen}{0.0391}}&{\cellcolor{lgreen}{0.0483}}&{\cellcolor{lgreen}{0.0474}}&{\cellcolor{lgreen}{0.0468}}&{\cellcolor{lgreen}{0.0471}}&{\cellcolor{lgreen}{0.0532}}\\ \hline
\hline
&\textbf{P@10}&\textbf{R@10}&\textbf{F1@10}&\textbf{NDCG@10}&\textbf{P@10}&\textbf{R@10}&\textbf{F1@10}&\textbf{NDCG@10}&\textbf{P@10}&\textbf{R@10}&\textbf{F1@10}&\textbf{NDCG@10}
\\ \hline
  \textbf{JoVA}&\cellcolor{lgreen}{0.2382}&0.1864&0.2092&0.2990&0.0272&0.0722&0.0395&0.0553&0.0406&0.0799&0.0538&0.0666\\ \hline
   \textbf{JoVA-H}&{0.2340}&{\cellcolor{lgreen}{0.1890}}&{\cellcolor{lgreen}{0.2115}}&{\cellcolor{lgreen}{0.3013}}&{\cellcolor{lgreen}{0.0281}}&{\cellcolor{lgreen}{0.0758}}&{\cellcolor{lgreen}{0.0401}}&{\cellcolor{lgreen}{0.0581}}&{\cellcolor{lgreen}{0.0409}}&{\cellcolor{lgreen}{0.0805}}&{\cellcolor{lgreen}{0.0542}}&{\cellcolor{lgreen}{0.0678}}\\ \hline
\end{tabular}
\end{center}
\padT
\caption{Performance of JoVA and JoVA-Hinge for various k, datasets, and metrics.  The gray shows the best value.}
\padT
 \label{tab:rankingLoss}
\end{table*}

\subsection{Evaluation Metrics}
 We utilize four commonly-used metrics to assess the quality of ranked list $\omega_u$ predicted for user $u$. \emph{Precision@k (P@K)} quantifies which fraction of u's recommended ranked list $\omega_u$ are $u$'s true preferred items:
\begin{equation}
P@k(\omega_u,I^*_u)= \frac{1}{k} \sum_{i=1}^k \ind{\omega_u(i) \in I^*_u},
\end{equation}
where $\ind{.}$ is the indicator function, $\omega_u(i)$ is the $i^{th}$ ranked item in $\omega_u$, and $I^*_u$ is $u$'s true preferred items in held-out data. Similarly, \emph{Recall@k (R@K)} measures which  fraction of $u$'s true preferred items $I^*_u$ are present in $u$'s recommended ranked list $\omega_u$:  
\begin{equation}
R@k(\omega_u,I^*_u)= \frac{1}{|I^*_u|} \sum_{i=1}^k \ind{\omega_u(i) \in I^*_u}.
\end{equation}

\emph {F1-score@k (F1@k)}, by computing the harmonic mean of the precision and recall, captures both of these metrics.  It reaches its maximum of 1, if both precision and recall are perfect (i.e., have value of 1):
\begin{equation}
    F1@k(\omega_u,I^*_u) = \frac{2 \cdot P@k(\omega_u,I^*_u) \cdot R@k(\omega_u,I^*_u)}{P@k(\omega_u,I^*_u) + R@k(\omega_u,I^*_u)}
\end{equation}
One shortcoming of P@k, R@k, and F1@K is giving the same importance to  all items ranked within the first k. To address this, NDCG@k gives higher weight to the higher ranked items:
\begin{equation*}
   NDCG@k(\omega_u,I^*_u) = \frac{1}{IDCG@k} \sum_{i=1}^k \frac{2^{\ind{\omega_u(i) \in I^*_u}} - 1}{\log_2(i+1)},
\end{equation*}
where $IDCG@k = \sum_{i=1}^{k}(1/\log_2 (i+1))$ normalizes NDCG with the maximum of 1.

We report the average of these metrics in our experiments, when the average is taken over all testing users. 

\subsection{Baselines} 
To evaluate the effectiveness of our methods, we compare them against various state-of-the-art recommendation methods. 
\vskip 1.5mm
\noindent \textbf{BPR} \cite{s_r+c_f+z_g+l_s+2012} optimizes a matrix factorization model with a pair-wise ranking loss. 

\vskip 1.5mm
\noindent \textbf{CDAE} \cite{w+y+d+c+2016}, by extending the denoising auto-encoder, assumes that observed ratings are corrupted
user’s preferences. 

\vskip 1.5mm
\noindent \textbf{Mult-VAE} \cite{l+d+k+r+2018} is a 
model with only one VAE, which deploys multinomial distribution for the output of the decoder. 

\vskip 1.5mm
\noindent \textbf{NCF} \cite{x_h+l_l+h_z+2017} 
learns user-item interaction function by combining both MF and multi-layer perceptrons with binary cross-entropy loss function. 

\vskip 1.5mm
\noindent  \textbf{JCA} \cite{z+z+w+j+2019} deploys two classical autoencoders for modeling users and items, and only uses hinge pairwise loss function. 

\vskip 1.5mm
\noindent \textbf{FAWMF} \cite{c+j+w+c+2020} is an adaptive weighted matrix factorization method based on a variational autoencoder.

For all baselines, we have used the implementations and optimal parameter settings reported by the original papers.

\subsection{Setup and Parameter Settings}
For learning all the models, we used the Adam optimizer with a learning rate of 0.003. For our models, as with \cite{z+z+w+j+2019}, we decomposed the training rating matrix into several small matrices, each of which is treated as a mini-batch. Each mini-batch size is set to encompass 1500 rows and 1500 columns. 

We run grid search on hyperparamerts and tested them against the validation sets.  We set $\lambda=0.15$ and $\alpha=0.01$ for all experiments, but picked $\beta$ individually for each dataset: $\beta = 0.001$ for Yelp, and $\beta = 0.01$ for both MovieLens and Pinterest. 
We randomly sampled one negative instance per a positive instance in each epoch. For each encoder and decoder, we had two hidden layers each with 320 dimensions and tanh activation functions while the sigmoid activation function was used for the output layers. For both VAEs, we set the dimension of the latent representation $d$ to 80.

\subsection{Performance Comparison}

\begin{figure*}[tb]
    \centering
    \begin{tabular}{@{\hspace{-7pt}}c@{\hspace{-1pt}}c@{\hspace{-1pt}}c}
       \includegraphics[width=0.33\textwidth]{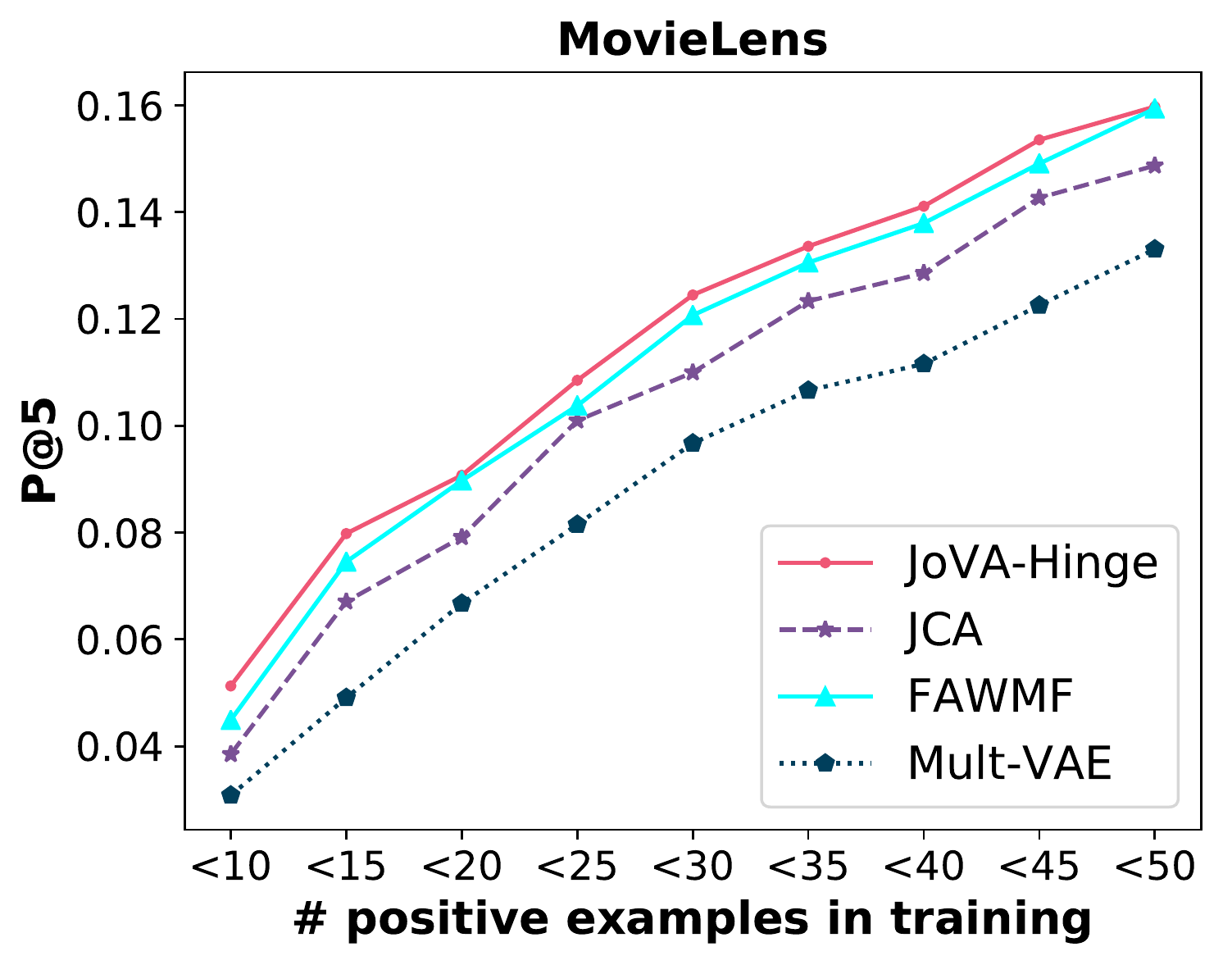}  &
       \includegraphics[width=0.33\textwidth]{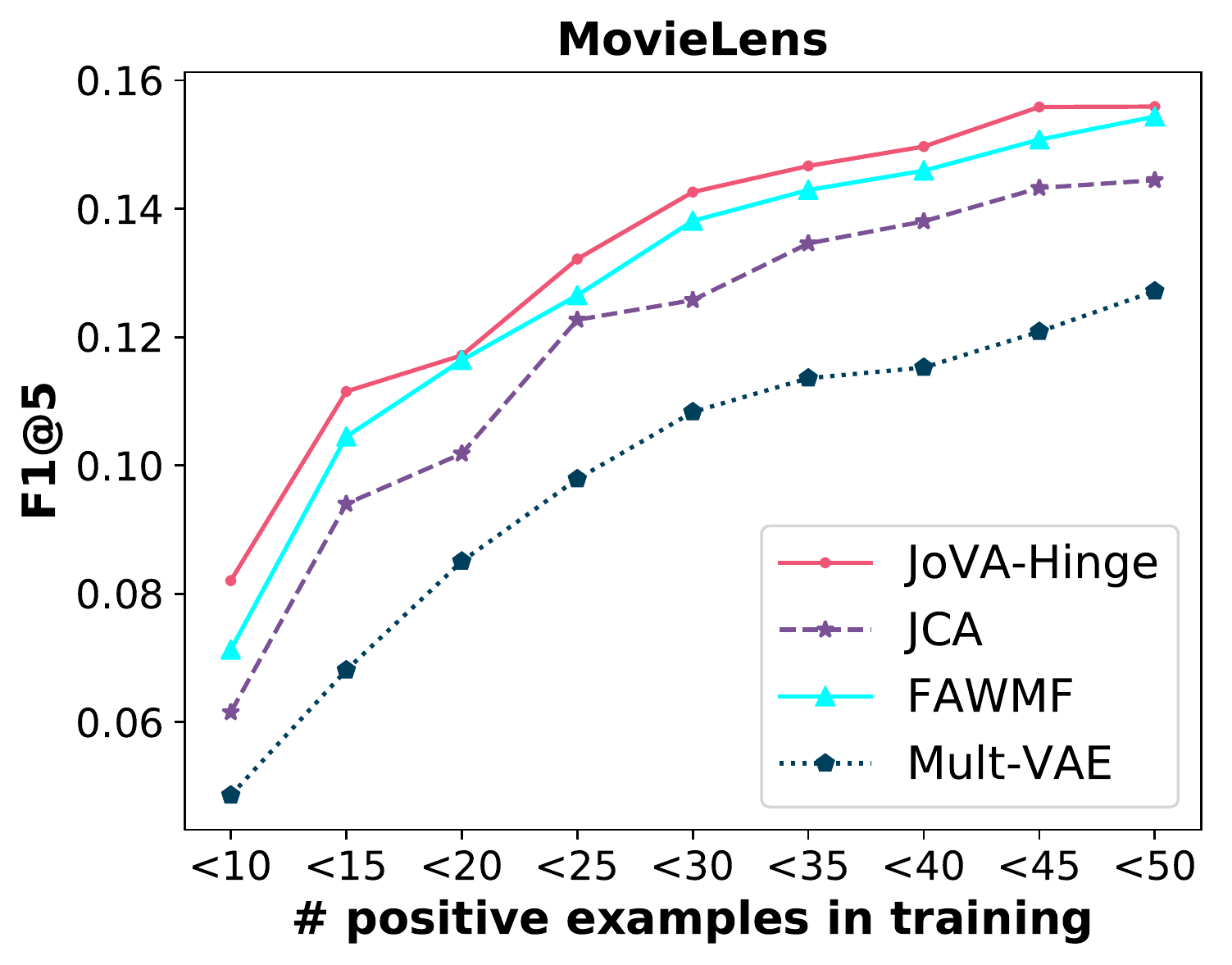} &
       \includegraphics[width=0.33\textwidth]{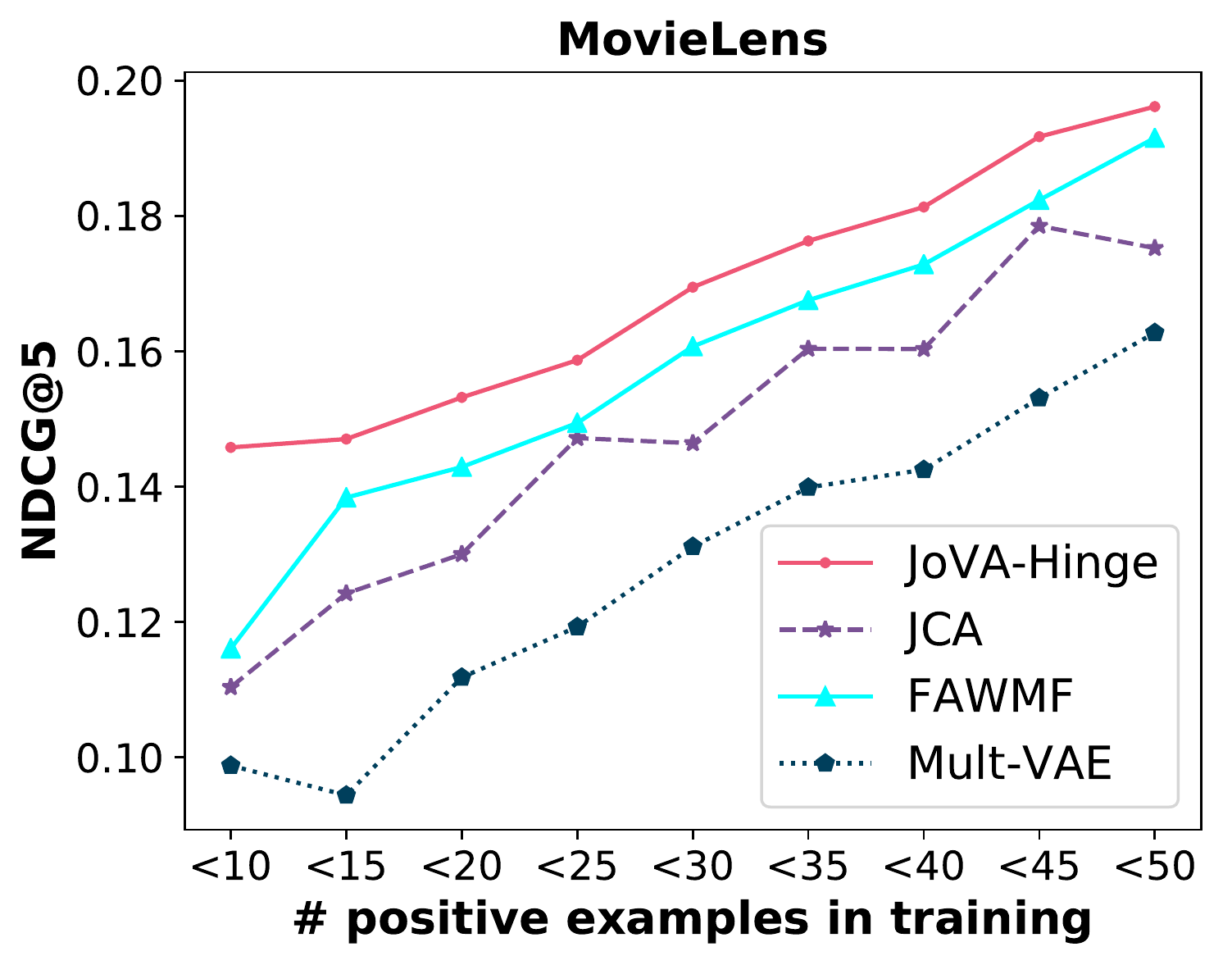}\\
       (a) Precision@5 &(b) F1-Score@5& (c) NDCG@5\\
    \end{tabular}
    \padT
    \caption{The average performance of users with the varying number of positive examples in their training data, MovieLens dataset, various evaluation metrics: (a) P@5, (b) F1@5, and (c) NDCG@5.}
    \label{fig:cold}
    \pad
\end{figure*}

We compare the performance of the top-k recommendation of our models and baselines with various $k \in \{1, 5, 10\}$ on different datasets. Figure \ref{fig:pre-rec-all} illustrates the performance of all methods under precision@k and recall@k for various $k$ and datasets. We first notice that neural network-based methods outperform the traditional ranking approach BPR, which suggests the effectiveness of non-linear features in learning the user-item interaction function. The JoVA-Hinge outperforms all methods (with a few small exceptions) on all datasets for both Precision@K and Recall@K. 

For more holistic performance analyses of all methods, Table \ref{tab:my-table} reports F1-Score and NDCG for all datasets and methods. Our JoVA-Hinge model outperforms others for F1 measure on three datasets and various k. Compared with the best baseline (JCA), F1-score@k is improved by up to 3.65\% in ML1M, 25.62\% in Yelp, and 33.33\% in Pinterest. For NDCG, JoVA-Hinge also outperforms others significantly in two datasets of Yelp and Pinterest. In Yelp, the mimimum improvement is 8.19\% (for $k=10$) and the maximum improvement is 10.86\% (for $k=1$). The JoVA-Hinge has even higher improvement for Pinterest with the mimimum of 21.72\% (for $k=10$) and the maximum of 34.82\% (for $k=1$).  For ML1M and NDCG, the performance of JoVA-Hinge is comparable to the performance of best baseline FAWMF. Cross-examination of Tables \ref{tab:PPer} and \ref{tab:my-table} suggest that our JoVA-H model significantly improves the accuracy of the state-of-the-art methods in terms of both F1 and NDCG for sparser datasets (i.e., Yelp and Pinterest). Our results also suggest that JoVA-Hinge offer more significant improvement for smaller $k$ (e.g., $k=1$ or $k=5$), which is of special practical interest for reducing cognitive burden on users, when the recommendation slate is small.

\subsection{Effect of Pairwise Ranking loss }
We aim to understand whether both variational encoders and pairwise loss function have contributed to the success of JoVA-Hinge. In other words, we are interested in assessing the effectiveness of the pairwise ranking loss in improving our model's accuracy. Thus, we compare the performance of JoVA-Hinge with JoVA in Table \ref{tab:rankingLoss} on three datasets and under four evaluation metrics. 

Our experiment illustrates that the pairwise ranking loss combined with VAE losses improves accuracy on almost every cases (except for p@1, p@10, and NDCG@1 on MovieLens). This finding suggests, by combining hinge loss function with VAEs, one can take advantage of capturing the relative distance between positive and negative items to learn more informative latent representations. We believe this successful marriage of VAEs and pairwise loss functions can be extended to other models built based on VAEs building blocks even in other applications  (e.g., vision, speech, etc.).

\subsection{Cold-Start and Data Sparsity}
Data sparsity and cold-start problems---dealing with users and items with few or no interactions---are practical challenges in recommender systems. We aim to understand how the accuracy of recommendation changes for users with a different number of user-item interactions (i.e., positive examples). We study the average accuracy of users with at most $L$ user-item interactions in training data while increasing $L$. This setting allows us to study not only cold-start users with small $L$ (e.g., $L=10$), but also how more availability of user-item interactions affect the accuracy of recommendation.     

Figure \ref{fig:cold} shows the performance of the top four methods of previous experiments (i.e., Mult-VAE, FAWMF, JCA, and JoVA-Hinge) under different metrics when $L$ increases.\footnote{The results for $k=1$ and $k=10$ were qualitatively similar,  which are not included due to space constraints.}  Unsurprisingly, the performance of all methods increases with more availability of user-item interactions. However, JoVA-Hinge outperforms other methods not only for users with the low number of user-item interactions (i.e., cold-start users), but also for well-established users. This suggests that the overall success of JoVA-Hinge is not limited to a specific class of users, and all users with various numbers of user-item interactions can benefit from its prediction power.

\section{Concluding Remarks and Future Work}
We have introduced joint variational autoencoder (JoVA) for top-k recommendation with implicit feedback.  JoVA, as an ensemble of two VAES, simultaneously and jointly learns user-user and item-item correlations. A variant of JoVA, referred to as JoVA-Hinge, includes pairwise ranking loss in addition to VAE's losses to specialize JoVA further for recommendation with implicit feedback. Our empirical experiments on three real-world datasets show that JoVA-Hinge advances the recommendation accuracy compared to a broad set of state-of-the-art methods, under various evaluation metrics. Additionally, our experiments demonstrate that the outperformance of JoVA-Hinge is across all users regardless of their numbers of observed interactions. 

Our JoVA model provides a solid framework for the broader investigation of the ensemble of VAEs equipped with pairwise ranking loss in recommender systems or possibly in other applications (e.g., vision, robotics, etc.). One can explore extending JoVA-Hinge to incorporate user and item features (e.g., descriptions, demographic information, etc.), side information (e.g., social networks), context (e.g., time, location, etc.), or non-stationary user preferences. 
\bibliography{main}

\end{document}